# Trainee Action Recognition through Interaction Analysis in CCATT Mixed-Reality Training

**Divya Mereddy, Marcos Quinones-Grueiro,**
**Ashwin T S, Eduardo Davalos, Gautam Biswas**
**Vanderbilt University, Nashville, TN**
divya.mereddy@vanderbilt.edu,
marcos.quinones.grueiro@vanderbilt.edu,
ashwin.tudur.sadashiva@vanderbilt.edu,
eduardo.davalos.anaya@vanderbilt.edu,
gautam.biswas@vanderbilt.edu

**Kent Etherton, Tyler Davis, Katelyn Kay,**
**Jill Lear**
**Air Force Research Laboratory, OH, USA**
kent.etherton.1@us.af.mil,
william.t.davis294.mil@health.mil ,
katelyn.kay.1@us.af.mil,
jill.d.lear.ctr@health.mil

**Benjamin S. Goldberg**
**US Army CCDC Soldier Center, Orlando, FL**
benjamin.s.goldberg.civ@mail.mil

## ABSTRACT

This study examines how Critical Care Air Transport Team (CCATT) members are trained using mixed-reality simulations that replicate the high-pressure conditions of aeromedical evacuation. Each team—a physician, nurse, and respiratory therapist—must stabilize severely injured soldiers by managing ventilators, IV pumps, and suction devices during flight. Proficient performance requires clinical expertise and cognitive skills, such as situational awareness, rapid decision-making, effective communication, and coordinated task management, all of which has be maintained under stress. Recent advances in simulation and multimodal data analytics enable more objective and comprehensive performance evaluation. In contrast, traditional instructor-led assessments are subjective and may overlook critical events, thereby limiting generalizability and consistency. However, AI-based automated and more objective evaluation metrics still demand human input to train the AI algorithms to assess complex team dynamics in the presence of environmental noise and the need for accurate re-identification in multi-person tracking. To address these challenges, we introduce a systematic, data-driven assessment framework that combines Cognitive Task Analysis (CTA) with Multimodal Learning Analytics (MMLA). We have developed a domain-specific CTA model for CCATT training and a vision-based action recognition pipeline using a fine-tuned Human–Object Interaction model, the Cascade Disentangling Network (CDN), to detect and track trainee–equipment interactions over time. These interactions automatically yield performance indicators (e.g., reaction time, task duration), which are mapped onto a hierarchical CTA model tailored to CCATT operations, enabling interpretable, domain-relevant performance evaluations.

## ABOUT THE AUTHORS

**Divya Mereddy** is a Ph.D. student and research assistant at Vanderbilt University in the Open-Ended Learning Environments (OELE) Lab. Her research focuses on using AI to evaluate team performance metrics in Critical Care Air Transport Team (CCATT) training for the U.S. Air Force. Through AI, she aims to provide data-driven insights and support instructors, students, and researchers in their roles, improving training effectiveness. Before her Ph.D., she worked in the data science industry, applying AI solutions across various fields. She earned her Master's degree from the University of Cincinnati. Her research interests include self-supervised and multimodal learning analytics.

**Dr. Marcos Quinones-Grueiro** is a research scientist at the Institute for Software-Integrated Systems and an adjunct professor in the Computer Science department at Vanderbilt University. His research interests are developing and applying foundational machine learning and reinforcement learning methods in important areas such as smart mobility, multimodal analytics in education, safety in aerospace operations, and energy management optimization. He has over 100 peer-reviewed publications, and his research has been funded by the Army, NASA, and NSF.

**Dr. Ashwin T S** is a Research Scientist at the Institute for Software-Integrated Systems in the Department of Computer Science at Vanderbilt University. His research focuses on computer vision, multimodal machine learning, affective computing, learning technologies, and human-computer interaction. He has authored over 50 peer-reviewed publications, including multiple best paper awards and nominations, and his research has received funding from the Technology Innovation Hub at IIT Bombay, the National Science Foundation, and the Army Research Laboratory.

**Eduardo Davalos** is a Graduate Student Research Assistant at the Institute for Software-Integrated Systems at Vanderbilt University. He investigates the intersection of eye-tracking, multimodal learning analytics, and artificial intelligence. His work emphasizes leveraging machine learning and large language models (LLMs) to improve how humans interact with and interpret data, especially in educational environments. He is dedicated to developing scalable analytics that deliver meaningful insights to enhance learning experiences. Eduardo's research seeks to bridge human cognition and AI-driven analysis, expanding how technology can support understanding and decision-making.

**Dr. Gautam Biswas** is a Cornelius Vanderbilt Professor of Engineering and a Professor of Computer Science and Computer Engineering at Vanderbilt University. He researches Intelligent Systems with a focus on creating intelligent, open-ended learning environments that adapt to students' learning performance and behaviors. He has developed innovative multimodal analytics to study students' learning behaviors in various simulation and augmented reality-based training environments. He has published over 700 refereed papers. He is an IEEE Life Fellow, a Fellow of the Asia Pacific Society for Computers in Education, and a Fellow of the Prognostics and Health Management Society.

**Dr. Kent Etherton** is a Research Psychologist at the Air Force Research Laboratory at Wright Patterson Air Force Base in Dayton, OH. He earned his Ph.D. in Industrial/Organizational and Human Factors Psychology from Wright State University in 2021. His research focuses on competency modeling, fidelity assessment, and skill acquisition.

**Maj William Davis** is a Critical Care Air Transport physician serving as the Director of the 59th Medical Wing En Route Care Research Center at Joint Base San Antonio, Texas. He conducts translational research across the continuum of combat casualty care, with a specific focus on en-route care. He has over 50 refereed publications.

**Ms. Katelyn Kay** is a Research Psychologist with the Air Force Research Laboratory, 711th Human Performance Wing. She received her M.S. in Human Factors from Embry-Riddle Aeronautical University in 2022. Her research focuses on identifying data requirements that improve the capture of human performance data.

**Jill Lear** is a Research Nurse with multiple years of experience reviewing medical records from critical care air transport team transports. She is a clinical instructor with experience in medical simulation.

**Dr. Benjamin Goldberg** works as a Senior Scientist at the U.S. Army CCDC Soldier Center, Simulation and Training Technology Center (STTC) in Orlando, FL. His research in Modeling & Simulation emphasizes deliberate competency development, adaptive experiential learning in simulation-based environments, and ways to use AI tools and methods to craft personalized learning experiences. Currently, he leads a research program that develops adaptive training solutions to support the Synthetic Training Environment. Dr. Goldberg is also the co-creator of the award-winning Generalized Intelligent Framework for Tutoring (GIFT) and holds a PhD from the University of Central Florida.

# Trainee Action Recognition through Interaction Analysis in CCATT Mixed-Reality Training

**Divya Mereddy, Marcos Quinones-Grueiro, Ashwin T S, Eduardo Davalos, Gautam Biswas**

**Vanderbilt University, Nashville, TN**

divya.mereddy@vanderbilt.edu, marcos.quinones.grueiro@vanderbilt.edu, ashwin.tudur.sadashiva@vanderbilt.edu, eduardo.davalos.anaya@vanderbilt.edu, gautam.biswas@vanderbilt.edu

**Kent Etherton, Tyler Davis, Katelyn Kay, Jill Lear**

**Air Force Research Laboratory, OH, USA**

kent.etherton.1@us.af.mil, william.t.davis294.mil@health.mil , katelyn.kay.1@us.af.mil, jill.d.lear.ctr@health.mil

**Benjamin S. Goldberg**

**US Army CCDC Soldier Center, Orlando, FL**

benjamin.s.goldberg.civ@mail.mil

## INTRODUCTION

Developing competency in critical care skills for high-stakes medical transport requires adequate training and assessment. The Critical Care Air Transport Team (CCATT) program prepares multidisciplinary teams—typically a physician, nurse, and respiratory therapist—to stabilize critically injured patients during aeromedical evacuation. Training takes place in high-fidelity simulations that replicate the operational complexity of in-flight care, utilizing manikins to model real-world scenarios. Trainees must integrate cognitive and psychomotor skills, emphasizing situational awareness, decision-making, and task execution while managing patients in environments with noise, vibration, limited space, and time pressure. Robust, objective performance assessment is essential for effective learning (Pierce et al., 2002; Tosuncuoglu et al., 2018). Traditional CCATT assessments rely on expert observations during simulated missions (Jernigan et al., 2016; Schulz et al., 2016). However, these evaluations are constrained by the observer's cognitive load, difficulty in recalling detailed behaviors, inter-rater variability, and the labor costs of sustaining expert presence across sessions (Davis et al., 2014).

To address these limitations, we are developing an automated assessment framework for CCATT that combines Cognitive Task Analysis (CTA) and Multimodal Learning Analytics (MMLA). We created a domain-specific CTA model that delineates core CCATT objectives—maintaining patient sedation, managing airways and breathing, and stabilizing patient conditions—and links them to key cognitive functions and observable trainee actions. CTA decomposes training into a hierarchy of goals, subgoals, and associated skill requirements, while multimodal data collection (including video and audio) enables comprehensive performance capture and analysis. Together, these components support automated, interpretable, data-driven metrics aligned with training objectives and enhance After Action Reviews (AARs). As an initial step, we developed a CCATT-specific trainee action recognition module within the MMLA system.

Existing action recognition methods, particularly those designed for traditional nursing or clinical training, do not generalize well to CCATT due to unique equipment and interaction dynamics. Moreover, CCATT requires recognition of fine-grained, temporally segmented human–object interactions under occlusion, low lighting, and shifting viewpoints. To meet these demands, we employ a three-stage deep learning model based on the Cascade Disentangling Network (CDN) to detect the start and end of trainee–equipment interactions. These segments are then used to compute objective performance metrics. The development of a domain-specific CTA model that formalizes CCATT training objectives and links them to observable trainee actions for structured, automated assessment.

This paper highlights two primary contributions of this work. They are:

1. The development of a domain-specific CTA model that formalizes CCATT training objectives and links them to observable trainee actions for structured, automated assessment.
2. The implementation of a vision-based, three-stage deep learning model to detect trainee-equipment interactions, enabling the extraction of objective performance metrics in complex training environments.

## BACKGROUND

### Automated Assessment of Simulation Environments

Simulation-based training environments offer controlled, immersive settings that allow trainees to develop cognitive, psychomotor, and behavioral skills. Despite their increasing use, automated assessment systems in these environments are still underexplored. Current assessment methods, especially in CCATT training, mainly depend on manual observation by instructors. Recent progress in fields like nursing and military training shows the potential of combining CTA models with MMLA pipelines to support structured, data-driven performance assessments (Vatral et al., 2021; Vatral et al., 2022a; Biswas et al., 2020; Di Mitri et al., 2019; Mirchi et al., 2020; Vatral et al., 2022b). In this method, a detailed CTA of the training scenario generates a hierarchy of cognitive and psychomotor tasks, which then helps analyze multimodal learner data. While this prior research highlights the potential to connect behavioral data with hierarchical task models, these approaches have not yet been applied to the CCATT field, which presents unique challenges because of its medical (critical care) and mobile nature.

### Action Recognition and Trainee Equipment Interaction Analysis

Research on action recognition in simulation-based training—particularly in nursing education—remains limited (Vatral et al., 2022a). Quantitative methods that capture trainee behavior and performance are especially scarce. Although the computer vision literature has made modest progress on action recognition for clinical training, most studies emphasize sensor-based systems (Ijaz et al., 2022; Sozo et al., 2022), while video-based approaches yield lower accuracy (Hu et al., 2023). Moreover, many models rely on human feature–based recognition and neglect critical contextual factors, such as the specific equipment with which trainees interact.

In contrast, computer vision offers advanced human–object interaction (HOI) models, e.g., HOICLIP, PVIC, and CDN, which demonstrate strong performance in detecting object-directed actions (Zhang et al., 2021; Zhang et al., 2023; Ning et al., 2023). HOI models aim to recognize <human, object, action> triplets from visual input, enabling structured interpretation of activities. Traditional HOI methods employ either two-stage pipelines (detect humans/objects, then pair them for interaction classification) or single-stage pipelines (jointly predict triplets end-to-end). Two-stage approaches often incur inefficient pairing and local feature bias, whereas single-stage methods struggle to balance localization and interaction reasoning within shared representations. The CDN model (Zhang et al., 2021) addresses these limitations by disentangling the tasks of human–object pair generation and interaction classification within a cascaded decoding structure, thereby improving triplet prediction efficiency and reducing ambiguity in the learned representations.

## CCATT LEARNING ENVIRONMENT

CCATT is a specialized medical unit within the U.S. Air Force tasked with stabilizing and transporting critically injured soldiers during aeromedical evacuation missions (Ingalls et al.). CCATT is carefully designed to replicate the operational, environmental, and cognitive demands faced by CCATT personnel during in-flight care. A typical CCATT team includes a physician, a nurse, and a respiratory therapist. The CCATT mixed-reality training is divided into two phases: *initial* and *advanced*. This study uses data from the initial CCATT simulation environment. The simulation for the initial phase includes two CCATT teams and four manikin-based patient beds, each equipped with essential critical care tools such as *mechanical ventilators* (MV), *intravenous* (IV) *infusion equipment* (pump, IV bags, IV tubes, IV hub), and *ProPaq* monitors.

Every team of three trainees manages one or two patients at a time, requiring high situation awareness, quick decision-making, and complex clinical procedures in a crowded, high-pressure setting. Multiple cameras are positioned throughout the simulation room to capture all trainees and equipment, creating a detailed dataset for visual learning analytics. The setup features both stationary and manually adjustable cameras, offering a wide range of video angles. However, camera coverage is inconsistent: some cameras offer a clear, continuous view of critical trainee–patient–equipment interactions, while others are limited by range, angle misalignment, or occlusion. Additional complexities in the environment affect data quality and interpretability. Trainees often occlude equipment and other trainees, resulting in fragmented visual information. Lighting conditions are deliberately varied to mimic realistic in-flight situations, such as low-light settings illuminated by headlamps, infrared grayscale video (for easier instructor observation), and RGB footage.

The dataset analyzed in this study includes 20 simulation sessions and recordings from two cameras in each session (1-1 and 1-4), as shown in Figure 1. It totals approximately 30 hours of video and 3 million frames. Although the recorded sessions captured a variety of medical activities, our study focuses on trainee interactions with three key medical devices: IV equipment, mechanical ventilators, and the ProPaq monitor. These devices are crucial to CCATT operations, as most clinical interventions are performed through them. Therefore, they are essential for assessing trainee behavior and competency. We refer to a CCATT trainee's interaction with equipment as '*valid_interaction*' and the absence of interaction as '*no_interaction*'.

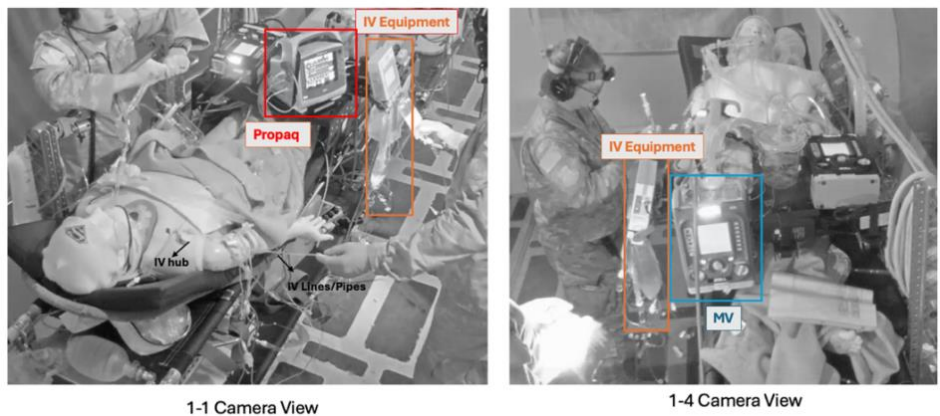

1-1 Camera View    1-4 Camera View

**Figure 1. CCATT Video Data Multi Camera Views (Source: AFRL, Dayton).**

**COGNITIVE TASK ANALYSIS MODEL AND PERFORMANCE METRICS FOR CCATT**

Through discussions with subject-matter experts and a review of the related literature (Vatral et al., 2021; Kinnebrew et al., 2017; Clark & Estes, 1996; Clark et al., 2008), we constructed a five-level hierarchical CTA model specific to CCATT operations (Figure 2). Our automated assessment system incorporates this multi-level CTA model to formalize CCATT's core clinical objectives and outline the procedural structure of transport missions. The five levels are described below.

**Level 1 – Operational Tasks**: Decomposes the mission objective into operational goals: (1) Keep the Patient Sedated, (2) Manage Airways and Breathing, and (3) Stabilize Patient Condition(s). Each operational task addresses distinct aspects of critical care and reflects real-time decisions trainees must make in flight. These tasks provide a structured approach to organizing complex responsibilities under time pressure.

**Level 2 & Level 3 – Cognitive Processes and Skills (STE-General tasks)**: At these levels, Level 1 goals are partitioned into cognitive processes and skills that represent more abstract, transferable behaviors (e.g., situational awareness, decision-making, task execution). Together, these constructs define the cognitive and psychomotor demands trainees must master to demonstrate adequate individual performance.
- *Situational Awareness*: Continuously monitor patient vitals, visual alarms, and equipment states. As shown in the CTA, situational awareness encompasses perception (e.g., noticing alarms), comprehension (e.g., interpreting data), and anticipation (e.g., forecasting patient instability).

- *Decision-Making*: Rapidly assess the situation and select interventions under uncertainty. Decision-making involves gathering, integrating, and analyzing information, as well as prioritizing clinical responses.
- *Task Execution:* Execute clinical actions—such as ventilator adjustments and IV line management—accurately and on time. Task execution includes medication preparation and manipulation of medical devices.

**Level 4 – Observable Actions or Events**: Operationalizes the integrated cognitive processes and skills into directly observable behaviors demonstrated in training. These include actions such as manipulating medical devices or responding to visual alarms. Observable events are critical because they can be consistently annotated, tracked, and aligned with multimodal data, bridging high-level cognitive intent and measurable trainee behavior. This layer most closely corresponds to performance metrics.

**Level 5 – Performance Metrics**: Maps observable actions to quantifiable indicators. Interpreted within the hierarchical CTA, these metrics provide insight into trainee proficiency, cognitive load, and readiness. Examples include fixation duration, alarm reaction time, and interaction start/end times, which serve as proxies for underlying cognitive competencies. Considered alongside higher-level goals, they enable data-driven evaluation and personalized feedback. All CCATT metrics are detailed in Table 1 and are supported by the MMLA pipeline.
.

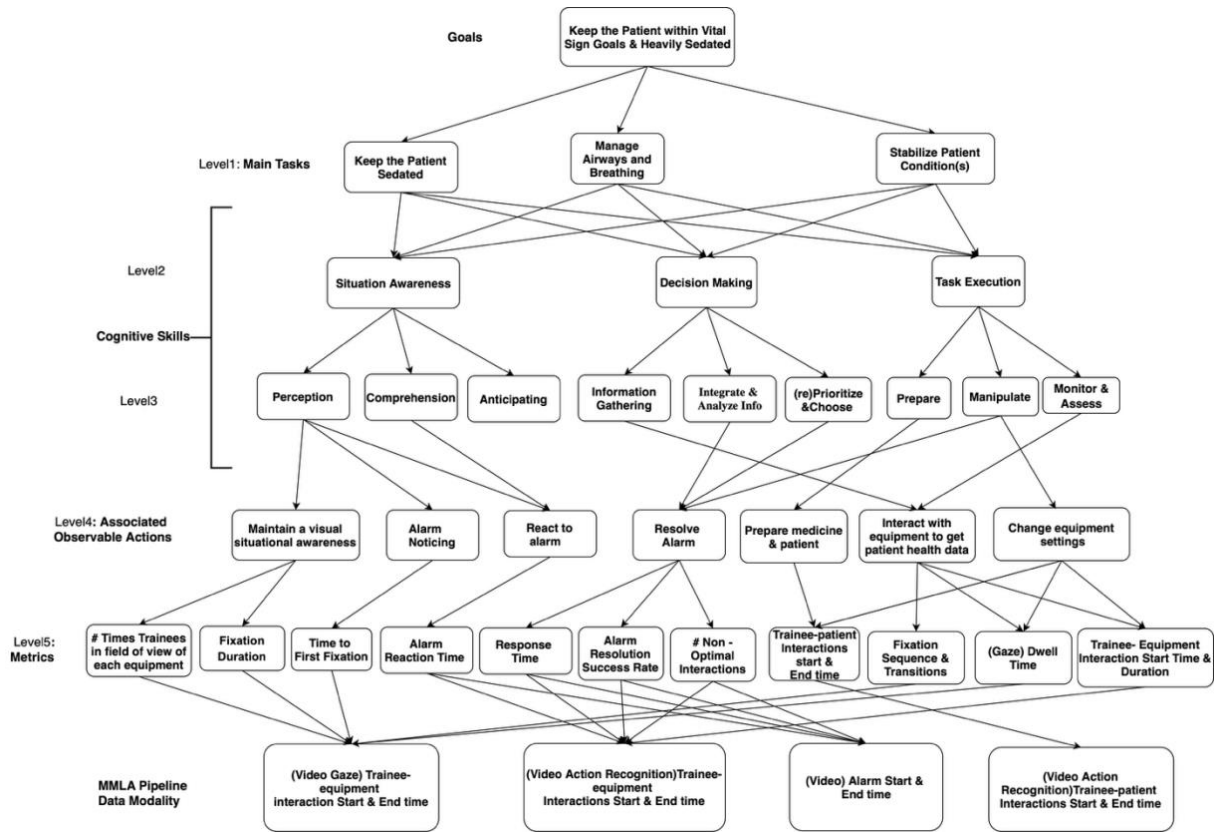

**Figure 2. CCATT Cognitive Task Model.**

**Table 1. Performance Metrics of CCATT Cognitive Task Model**

| Metric | Description |
|---|---|
| # Times Trainee in the Field of View of Equipment | Counts how often a trainee appears within the field of view of critical equipment (IV, ventilator, monitor). |
| Fixation Duration | Duration of sustained gaze fixations per minute on the equipment. |

| Time to First Fixation | Elapsed time between alarm onset and the first gaze fixation on the equipment. |
|---|---|
| Alarm Reaction Time (Physical) | Time between alarm onset and the trainee's first physical interaction with associated equipment. |
| Response Time | Total duration from alarm onset to alarm resolution. |
| Alarm Resolution Success Rate | Proportion of equipment interactions that successfully resolve alarms. |
| # Non-Optimal Interactions | Number of interactions that lead to false alarms or ineffective responses. |
| Trainee–Patient Interaction Start & End Times | Timestamps marking the beginning and end of physical contact with the patient. |
| Fixation Sequence & Transitions | Patterns of gaze movement between equipment areas (e.g., from alarm to vital signs monitor). |
| Gaze Dwell Time | Duration of uninterrupted gaze on area of interest (infer depth of processing). |
| Trainee Equipment Interaction Start & End Times | Timestamps denoting when the trainee begins and ends interaction with a CCATT equipment. |

**THREE-STAGE HUMAN OBJECT INTERACTION (HOI) DEEP LEARNING MODEL**

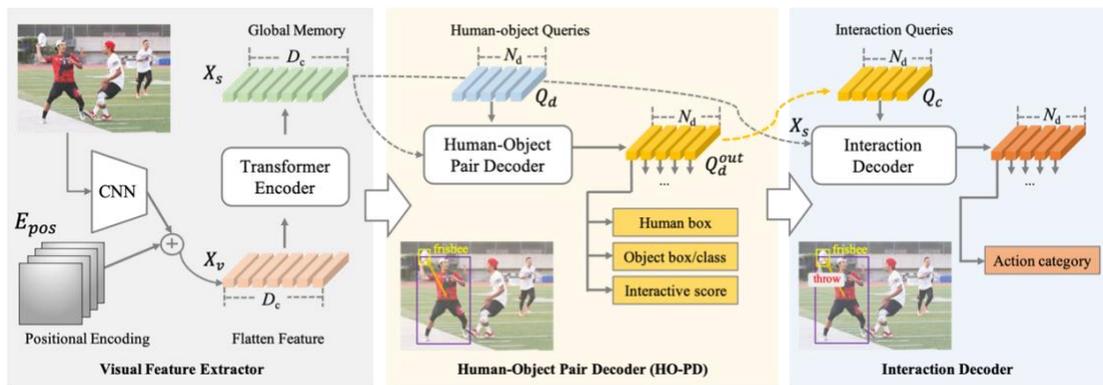

**Figure 3. Cascade Disentangling Network (CDN) Model Architecture (Zhang et al., 2021).**

Unlike traditional action recognition models, which aim to assign a single global activity label to an individual, HOI detection explicitly models triplets consisting of a human, an object, and their interaction. This approach provides a more semantically rich and interpretable framework, particularly in environments like CCATT, where most actions involve objects. In CCATT, actions are mediated through equipment. The environment presents significant challenges for HOI modeling due to its highly dynamic and densely populated nature, where multiple trainees interact with several critical care devices at once. Interactions often involve subtle physical cues, such as touching equipment versus merely standing near it with hands close without actually engaging; the confined and crowded setting further complicates this. Additionally, the equipment appears similar and is positioned near other devices and trainees. These factors necessitate a model that can distinguish between object localization and interaction classification. The CDN model (Zhang et al., 2021) introduces a three-stage decoupled architecture, as shown in Figure 3. The CDN is specifically designed to clearly differentiate context modeling, entity localization, and interaction classification, effectively resolving the representational conflicts seen in earlier methods. The CDN framework breaks down the HOI detection task into three stages.

***Stage 1. Feature extraction through Scene-Level Contextualization:*** The first stage employs a transformer encoder adapted from DETR (DEtection TRansformer), integrating CNN-based local feature extraction with transformer-based global context modeling. A convolutional backbone (e.g., ResNet) extracts fine-grained local visual features, capturing subtle patterns such as hand gestures, finger positions, and facial orientation, which are critical for identifying nuanced equipment interactions. These local features are then processed by transformer layers that model long-range spatial dependencies across the scene, encoding broader contextual cues such as room layout, inter-object proximity, and human–object proximity. This dual modeling enables the network to jointly reason over low-level visual cues and high-level scene semantics—both essential in CCATT environments, where micro-movements and situational context jointly indicate actions.

***Stage 2. Human-Object Pair Decoder (HO-PD):*** The HO-PD identifies interactive human–object pairs. Each query is decoded into a human bounding box, an object bounding box, and an object class. This stage also outputs a set of human–object candidates and retains their final query embeddings Qdout, which are passed to the next stage. The HO-PD and the subsequent Interaction Decoder are built on the same transformer-based decoder architecture (following DETR design) but use independent learnable parameters and task-specific prediction heads. Each decoder comprises stacked transformer layers, each with a self-attention module (to model inter-query relationships) and a multi-head co-attention module (to relate queries to encoder-derived visual features). Feed-forward networks attached to each query perform task-specific predictions, such as bounding boxes and object classes.

***Stage 3. Interaction Decoder***: The Interaction Decoder consumes enriched HOI queries (Qdout) from the HO-PD as input queries. While it shares the transformer design of the HO-PD, it employs sigmoid heads tailored for verb classification, enabling multi-label verb classification for each human–object pair based on the decoded pair representations.

The decoding process explicitly separates object-level reasoning from interaction classification, allowing the model to attend to subtask-relevant features and reducing entanglement between object and action predictions. As a foundational baseline for automated interaction recognition in the CCATT domain, we applied the small CDN variant pretrained on HICO-DET (Humans Interacting with Common Objects – DETection) directly to CCATT simulation videos in a zero-shot setting. This served as our base model to assess the transferability of HOI representations to the specialized medical domain of critical care air transport.

## CDN Model for CCATT HOI detection

Although the pretrained CDN model was never exposed to CCATT-specific equipment categories, it learned to associate human pose features—such as extended arms or grasping gestures—with nearby object regions. Verb predictions in CDN are mainly influenced by pose features, which generalize well from the HICO domain to our CCATT domain. When a valid trainee–equipment interaction appears in the frame, with similar physical movements like a trainee reaching toward equipment, the model assigns higher confidence scores to verbs like 'hold' or 'carry'. At the same time, due to learned spatial patterns, the model pays more attention to objects near the hands and identifies them. However, since the model was not trained on CCATT objects, it often assigns incorrect labels and sometimes draws inaccurate bounding boxes. Because CDN predicts human–object interactions jointly, an increase in verb confidence often slightly raises object confidence as well. Nonetheless, the overall HOI confidence remains low, as neither component is firmly grounded in the CCATT domain. This behavior is typical of zero-shot HOI transfer, where verb generalization outpaces object recognition because of the domain-specific gap in visual representations.

To leverage accurate verb predictions even when object detection is low-confidence or misclassified, we implemented a bounding-box similarity strategy to map predicted boxes with non-relevant class labels to CCATT equipment. Predicted object bounding boxes were matched to predefined CCATT equipment regions based on spatial overlap, using the Intersection over Union (IoU) metric. If a predicted box had a sufficiently high IoU with a known equipment region, it was considered to represent that equipment, regardless of the model's predicted class label. This approach allowed us to derive potential CCATT interactions from pretrained HOI triplets. The outputs were refined further as follows:

- Verb Class Filtering: Only interactions involving the verbs hold, carry, watch, or no interaction were kept, as they are closely aligned with CCATT trainee–equipment interaction behaviors. The mapping is defined as: *"valid_interaction"* → hold or carry; *"no_interaction"* → watch or no interaction.
- Final HOI Assignment: For each object, the highest-scoring CCATT-relevant HOI was selected to represent the interaction label. To minimize misclassification and reduce prediction noise caused by multiple CCATT objects, we cropped each frame as much as possible to retain only the relevant equipment and interaction with the trainee.

**Fine-Tuning the CDN Architecture for the CCATT Environment**

Adapting the pretrained CDN model to the CCATT environment proved difficult. It had trouble detecting CCATT equipment, such as IV equipment, MV, and ProPaq, and showed limited ability to recognize verbs relevant to CCATT. Although some HICO verbs, like hold, carry, and watch, showed semantic similarities to CCATT-relevant actions, the model lacked enough contextual understanding to distinguish subtle but important behavioral cues that separate *valid_interaction* from *no_interaction* in CCATT scenarios. This lack of contextual understanding made it hard to produce confident predictions. To adapt the pre-trained CDN-HICO model to the CCATT domain, we started fine-tuning the model. The steps for fine-tuning are outlined below.

***Step 1. Semi-automated data labeling process for training data annotations:*** Since model fine-tuning requires static HOI labels for training, we developed a semi-automated labeling pipeline with manual validation. We began by extracting individual frames from multi-camera CCATT training videos (Camera 1-1 and 1-4 views). Next, we used a pretrained CDN model, adapted to the CCATT environment as described earlier, to generate initial interaction predictions. Frames with high-confidence interactions were kept, and the top-scoring interactions within these frames were labeled as "*valid_interaction*". Low-confidence predictions might represent either valid object detections with no interaction ("*no_interaction*") or model uncertainty, leading to incorrect object predictions. Accurate labeling of all interactions was crucial to reduce annotation ambiguity and enhance fine-tuning. To address this, we added a secondary verification step using MediaPipe hand-skeleton estimations (Lugaresi et al., 2019). We re-evaluated frames: overlapping hand skeletons with objects confirmed true interactions, while the absence of overlap—even at lower skeleton confidence thresholds—was regarded as likely "*no_interaction*" cases. This two-step validation improved accuracy for both positive and negative interaction labels. To remove redundant human detections generated by the pretrained CDN model, we applied clustering-based non-maximum suppression (NMS) to eliminate duplicate human-object pairs. We also used a YOLO-based human detection module (Redmon et al., 2016) to recover missed human instances that CDN failed to detect, ensuring annotation completeness. Finally, we randomly selected 5 images for manual verb correction to address residual labeling errors. The dataset was then augmented with photometric transformations (such as lighting variation and Gaussian blur) to generate a total of 20,000 labeled images, which were used for fine-tuning.

***Step 2. Expert annotation process for test data creation***: To establish a dependable benchmark for evaluating our HOI recognition model, we teamed up with a certified CCATT training instructor to develop expert-annotated test data. The instructor temporally annotated training videos, marking intervals where interactions between trainees and critical equipment (IV equipment, MV, Propaq) took place. Due to the temporal nature of these annotations, the labeling approach deliberately allows for brief non-interactive frames within an ongoing interaction sequence. This decision aligns with our system's main goal: to support temporally grounded evaluation of trainee behavior rather than frame-by-frame precision. These temporally consistent labels function as the gold standard in our evaluation process.

***Step 3. Fine-tuning the stages of the CDN model for the CCATT environment***: For fine-tuning, we used a two-stage approach that tackled the specific challenges of object detection and verb classification in the CCATT simulation environment. Instead of adding entirely new classes, which would require architectural adjustments and training new parameters from scratch, we mapped CCATT-specific interactions and equipment to semantically similar HICO categories. For example, "*valid_interaction*" was mapped to "*hold*," and "*no_interaction*" was aligned with the HICO "*no_interaction*" verb. We also assigned equipment bounding boxes to similar object IDs in HICO. This mapping was based on how often the related object and verb categories appeared in the pre-trained model results, as well as the semantic features of the objects, verbs, and their interactions.

In the first stage of fine-tuning, we trained the entire CDN model, including the visual feature extractor (ResNet-50 + Transformer encoder) and both decoders, in an end-to-end manner. This joint optimization of the encoders and decoders helped align visual feature representations with the task-specific goals of detecting CCATT equipment and classifying trainee interactions. This stage resulted in a moderate reduction in verb classification loss and a significant improvement in object detection accuracy. However, due to the class imbalance in our domain, where "*no_interaction*" is overrepresented, the model showed suboptimal verb recognition accuracy for the minority class, "*hold*" (*valid_interaction*). In the second stage, following CDN (Zhang et al., 2021) decoupling training strategy, we conducted partial architecture training. The visual feature extractor was frozen, and only the two decoders were fine-tuned. This strategic isolation allowed the HO-PD to retain previously learned spatial localization priors while enabling the Interaction Decoder to specialize in refining verb classification. This decoder-focused training phase led to significantly improved verb scores, reduced bias toward the positive class, and enhanced confidence distributions for CCATT interactions.

Human-object interaction scores were computed as the product of object detection confidence and verb classification probability, often exhibiting sharp and erratic fluctuations from frame to frame. To further stabilize predictions at the frame level and align model behavior with the physical constraints of real-world interactions, we incorporated a temporal Gaussian smoothing. After smoothing across the temporal sequence of predictions, we applied a threshold-based cutoff method to segment active interaction intervals. The standard deviation (σ) of the Gaussian kernel and the segmentation threshold were empirically selected via grid search, using validation data CCATT temporal expert annotations.

**RESULTS**

Based on (Zhang et al., 2021), we selected our model hyperparameters, optimizer, and learning rate settings. We used session-independent cross-validation, dividing the data as 70% for training, 15% for testing, and 15% for validation. For testing, we randomly sampled data snippets without complete occlusions from the test dataset to evaluate the model.

**Evaluation Metrics for Trainee- Equipment Interaction Detection Model**

To evaluate model performance, we used the frame-level F1 score, which is appropriate for our dataset due to its significant class imbalance, with non-interaction labels greatly outnumbering interaction cases. The F1 score is the harmonic mean of precision and recall (see Equation 1), balancing false positive and false negative rates. Precision measures the proportion of correct interaction predictions among all predicted interactions, while recall tracks the number of actual interactions correctly identified. This makes F1 a reliable metric for assessing both frequent and rare interaction classes. We calculated training F1 scores on static image frames for verb classification. For testing, we applied temporal smoothing to model predictions before evaluation, capturing interaction continuity over time. F1 scores were then computed on smoothed predictions to reflect realistic temporal behavior in CCATT.

$$F1 = \frac{2 \cdot Precision \cdot Recall}{Precision + Recall} \times 100; \quad (1) \qquad Precision = \frac{TP}{TP + FP} \qquad Recall = \frac{TP}{TP + FN}$$

Here, TP: True Positives (correctly predicted positives), FP: False Positives (incorrectly predicted positives), FN: False Negatives (missed actual positives), TN: True Negatives (correctly predicted negatives).

**Results for Trainee- Equipment Interaction Detection Model**

Model performance was evaluated using frame-level macro F1 scores after temporal smoothing (in percent). As shown in the comparison of results for the pre-trained CDN model versus the fine-tuned model in Table 2, the fine-tuned model demonstrated balanced performance for all equipment interactions, achieving an 87.2 macro F1 score. In addition, the fine-tuned model, compared to the pretrained model, showed clear gains in F1 score for all equipment: an absolute increase of 42.7 points for IV, 37.4 for MV, and 35.5 for ProPaq.

**Table 2. Pretrained vs Fine-tuned Model**

|  | Pretrained F1 | Fine-tuned F1 |
|---|---|---|
| **IV Equipment** | 52 | 94.7 |
| **MV** | 47 | 84.4 |
| **ProPaq** | 47 | 82.5 |
| **Avg.** | 48.6 | 87.2 |

**Domain-Specific Action Recognition Model Evaluation**

To evaluate the contextual robustness and predictive reliability of our fine- tuned HOI detection model in the CCATT training environment, we calculated the following model evaluation metrics:

1.  *Temporal Overlap Ratio* (Ground Truth vs. Prediction) measures the fraction of the ground truth(GT) interaction duration that is accurately covered by the predicted interval.

2. *Falsely Predicted Interactions Count* quantifies the number of interaction intervals that were falsely predicted (not present in GT), reflecting the model's effectiveness in avoiding false positives.

3. *False Interaction Prediction Duration* (% of Total Predicted Duration) captures the proportion of total predicted interaction time that was inaccurate. It provides insight into the quality of the model's interaction score predictions.

4. *Model Start Time Latency* measures the average delay between the start of ground truth interaction events and the corresponding predicted interaction start times, calculated only for overlapping GT-prediction pairs.

These metrics are designed to assess the model's predictive accuracy, temporal consistency, and suitability for use in high-stakes simulation environments. Our experiments using domain-specific metrics show that finely tuned models consistently outperform pretrained models across all devices (Table 3). The performance differences are visually confirmed in the score plots (Figure 4). For IV equipment, the fine-tuned model achieved a 98.02% overlap ratio, with only one false interaction and no start-time latency, while the pretrained model failed to capture any valid interactions, indicating a complete failure to detect meaningful interactions. For the MV, the fine-tuned model achieved an 86.6% overlap ratio, whereas the pretrained model showed low overlap (88.2%) and poor interpretability. For ProPaq, the fine-tuned model reached a 78.53% overlap, while the pretrained model again failed to identify any valid interactions and predicted very low interaction scores, as shown in Figure 4. The start time latency values are not reported for the pretrained model, as its predictions were misaligned with the ground truth.

**Table 3. Domain-Specific Evaluation Results for Trainee Action Recognition Model**

| Equipment | Model Type | Overlap Ratio | Avg. Falsely Predicted Interactions Count | Model Start Time Latency (sec) | False Interaction Prediction Duration |
|---|---|---|---|---|---|
| **IV Equipment** | Finetuned | 98.02% | 1.00 | 0.0 | 2.91% |
| | Pretrained | 0.000 | 0.00 | - | 0.0% |
| **MV** | Finetuned | 86.60% | 3.00 | 0.28 | 17.49% |
| | Pretrained | 8.2% | 4.00 | - | 65.2% |
| **ProPaq** | Finetuned | 78.53% | 1.00 | 0.28 | 14.16% |
| | Pretrained | 0.000 | 0.00 | - | 0.0% |

**DISCUSSION AND FUTURE WORK**

The action recognition model accurately identified trainee–equipment interactions and their start and end times (see Figure 4). A key finding is the model's robustness and adaptability: by treating humans, equipment, and interactions as separate components, the three-stage design generalizes across environments and supports structured fine-tuning. Using the model's temporal predictions, we compute real-time clinical performance metrics. Among these, reaction time—defined as the delay between an equipment alarm and the predicted onset of trainee interaction—is a key indicator of situational awareness and other non-technical skills, reflecting how promptly a trainee perceives and responds to critical events. Temporal reasoning over predictions enables precise, automated measurement of reactivity. Beyond reaction time, the model supports metrics related to task comprehension, including interaction duration and the frequency of valid or ineffective interactions.

Model evaluation emphasized performance under occlusion and temporal blur, conditions that are common in CCATT scenarios. When the pre-trained version of the CDN model was evaluated on a subset of frames without partial occlusion or blur, it exhibited considerably higher performance than reported in the overall results. The reported pretrained metrics are derived from full-length videos that include partial occlusions, motion-induced blur, and reduced visual clarity. During fine-tuning, we incorporated examples with occlusion and lighting variations to enable the model to learn robust representations for localization and interaction recognition under degraded visual conditions.

Future work will extend the pipeline to include additional CCATT-relevant metrics defined in the Cognitive Task Analysis (CTA) model, such as interaction duration, task handover latency, action sequencing, and multitasking indicators. Although the system is robust to partial occlusions, it does not address complete occlusions where the body or equipment is entirely out of view; future enhancements will target occlusion robustness through temporal modeling and multi-view integration.

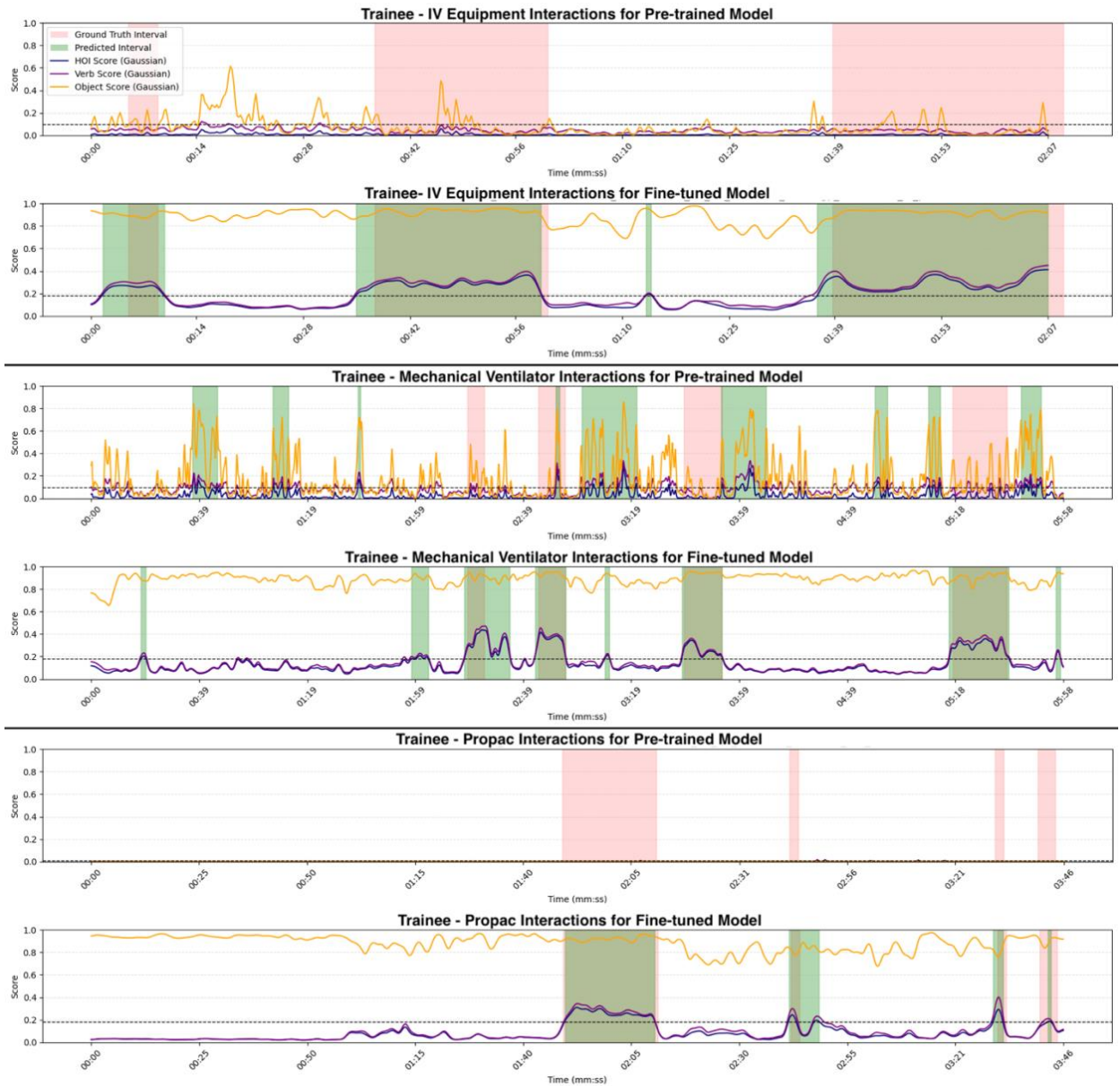

**Figure 4. Trainee-equipment interaction detection for pre-trained and fine-tuned models.**
(Trainee- ProPaq interaction scores for pretrained model are very low and not visible on 0-1.0 scale)

## SUMMARY AND CONCLUSIONS

This study marks an important advancement toward creating an automated trainee assessment system for high-stakes medical simulation environments like CCATT. Built on a Cognitive Task Analysis (CTA) framework and incorporating Multimodal Learning Analytics (MMLA) principles, we introduce a domain-specific CTA model tailored for the CCATT setting and develop a key component of the MMLA pipeline—a trainee action recognition system based on human-object interaction (HOI) models. The proposed model performs well in identifying equipment-specific actions, even under difficult conditions such as semi-occlusions and visual blur. Quantitative evaluation against expert annotations shows consistent temporal alignment, with average overlap ratios exceeding 87% across different equipment types. Beyond action detection, this work illustrates how observable CCATT domain-specific trainee evaluation metrics can be derived from MMLA outputs and aligned with low-level constructs in the CCATT-specific CTA. The combined CTA and MMLA system, supported by the action prediction model, offers a framework that is both interpretable and directly applicable.

**ACKNOWLEDGEMENTS**

The US Army CCDC Soldier Center Award (# W912CG2220001) has partially funded this research and development work. The views in this paper do not represent the position or policy of the United States Government, and no official endorsement should be inferred.

Distribution is unlimited.

Case Number: AFRL-2025-3176, 27 June 2025.